\documentclass{article}

\usepackage{microtype}
\usepackage{graphicx}
\usepackage{subcaption}
\usepackage{booktabs} 

\usepackage{amsmath}
\usepackage{amssymb}
\usepackage{amsthm}
\usepackage{mathtools}
\usepackage{caption}
\usepackage{multirow}

\newtheorem{prop}{Proposition}
\newtheorem{lem}{Lemma}
\newtheorem{thm}{Theorem}

\DeclareMathOperator*{\argmin}{argmin}

\usepackage{hyperref}

\usepackage[accepted]{icml2021}

\icmltitlerunning{Policy Evaluation when Causality is Uncertain}

\begin{document}

\twocolumn[
\icmltitle{Model-Free and Model-Based Policy Evaluation\\
            when Causality is Uncertain}




\begin{icmlauthorlist}
\icmlauthor{David Bruns-Smith}{ucb}
\end{icmlauthorlist}

\icmlaffiliation{ucb}{Department of Electrical Engineering and Computer Sciences,  University of California, Berkeley, USA}

\icmlcorrespondingauthor{David Bruns-Smith}{bruns-smith@berkeley.edu}

\icmlkeywords{Reinforcement Learning, Markov Decision Process, Causal Inference, POMDP}

\vskip 0.3in]



\printAffiliationsAndNotice{} 

\begin{abstract}
When decision-makers can directly intervene, policy evaluation algorithms give valid causal estimates. In off-policy evaluation (OPE), there may exist unobserved variables that both impact the dynamics and are used by the unknown behavior policy. These ``confounders'' will introduce spurious correlations and naive estimates for a new policy will be biased. We develop worst-case bounds to assess sensitivity to these unobserved confounders in finite horizons when confounders are drawn iid each period. We demonstrate that a model-based approach with robust MDPs gives sharper lower bounds by exploiting domain knowledge about the dynamics. Finally, we show that when unobserved confounders are persistent over time, OPE is far more difficult and existing techniques produce extremely conservative bounds.


\end{abstract}

\section{Introduction}
\label{intro}

Due to cost, feasibility, or safety concerns, practitioners often need to evaluate a sequential decision-making strategy using only previously-collected observational data. In reinforcement learning (RL), this problem is called off-policy policy evaluation (OPE). When the policy used to collect the data is unknown, there might exist unobserved variables correlated with both the policy and the outcomes. In this case, the causal effect of future interventions is unidentified and naive estimates for a new policy will be biased.

What kind of so-called unobserved confounders arise in Markov decision processes (MDPs)? Unobserved variables of interest in a medical setting are almost always highly persistent. For example, consider electronic medical records that do not document socio-economic status. A patient's socio-economic status is unlikely to change between visits to the hospital. In macroeconomics on the other hand, unobserved shocks are often assumed to be drawn iid every period. Consider the Federal Reserve Board adjusting monetary policy in response to oil price shocks. Events like earthquakes in oil fields might reasonably be assumed to occur independently across quarters.

Recent work develops OPE methods that are robust to unobserved confounding \cite{namkoong2020off, kallus2020confounding}. Given an observational data set and a hypothetical confounder, these methods adapt importance sampling approaches to calculate worst-case estimates for the value of a new policy. A practitioner can assess the sensitivity of their results to unobserved variables by increasing the strength of confounding and computing how quickly the worst-case bounds degrade.

However, the existing literature arrives at radically different conclusions. \cite{kallus2020confounding} - henceforth KZ - finds that it is possible to efficiently construct non-conservative bounds in the infinite horizon setting. On the other hand, \cite{namkoong2020off} - henceforth NKYB - only finds non-trivial bounds when confounding is restricted to a single time step. Furthermore, both approaches find the finite horizon case with confounding at each step to be computationally intractable. 

The natural questions are: 1) what is responsible for the substantial gap between the conservativeness of the existing bounds? and 2) how can we compute tractable lower bounds for the finite horizon case?

\textbf{Summary of our Results:}

We identify a key assumption under which it is possible to obtain sharp lower bounds on the expected value in a confounded MDP, even as the horizon grows. When the unobserved confounding variables are drawn iid each period, the marginal dynamics over the observed state themselves form an MDP. In this case, OPE methods can be applied to the marginal MDP after appropriate adjustments for confounding. Such an assumption is made in KZ.

But if the unobserved state might be persistent over time, the problem is a genuine partially-observed MDP (POMDP). Marginal transition probabilities for the observed state will not be Markovian in general. Medical applications, which frequently feature persistent unobserved variables, fall under this category. As a result, existing bounds that target this setting, such as NKYB,  are more conservative. In this paper, we focus on the case where the marginal problem is an MDP and demonstrate enormous performance differences compared to setting with persistent unobservables.

We derive an expression for the bias of common estimands under confounding in the marginal MDP setting. We show how to express OPE ``direct methods'' in this form. Then we demonstrate how to adapt direct methods to give worst-case bounds in the finite horizon case. Our method is sufficiently generic that any approach which regresses a function against states and actions can be plugged into our framework to get bounds.

Finally, we show that model-based OPE methods provide sharper lower bounds on the value function. We can compute these bounds in a computationally efficient way by combining techniques from the robust MDP literature with sensitivity models from causal inference. A model-based approach provides a natural way for domain experts to provide guidance on reasonable limits for the strength of confounding on outcomes. We evaluate our methods with existing OPE benchmarks.

\section{Related Work}
\label{related}

\textbf{Off-policy evaluation} 
There are several classes of popular OPE algorithms. \cite{voloshin2019empirical} provides a summary and empirically compares their performance. These classes include: importance sampling (IS) \cite{precup2000eligibility, hanna2019importance}, model-free direct methods like Fitted Q-Evaluation \cite{le2019batch}, model-based methods \cite{paduraru2012off, gottesman2019combining}, and hybrid methods \cite{thomas2016data, jiang2016doubly, kallus2020double}. \cite{voloshin2019empirical} shows that, typically, either simple methods like FQE or hybrid methods have the best performance in practice.

Recently, a variety of marginalized importance sampling (MIS) methods \cite{liu2018breaking, uehara2020minimax, nachum2020reinforcement} have been developed, which have the potential to solve the poor empirical performance of standard IS. This approach is adopted by KZ.

\textbf{Causal inference and sensitivity analysis}

Estimating the causal effect of a treatment on some outcome is the object of study in causal inference \cite{hernan2010causal, imbens2015causal, pearl2009causal}. The line of work on dynamic treatment regimes \cite{murphy2003optimal, laber2014dynamic} is the most relevant to RL. Work in this area frequently assumes an unconfoundedness condition, which guarantees that the causal effect of a treatment is identified. For example, unconfoundedness will hold if the data come from a randomized control trial.

If unconfoundedness might be violated, then a researcher can assess the robustness of their causal estimates via sensitivity analysis \cite{rosenbaum2002overt, franks2019flexible}. In recent work, \cite{yadlowsky2018bounds, kallus2019interval} give bounds for treatment effects subject to a sensitivity model. Other work develops bounds for the effectiveness of a single-step policy in the presence of unobserved confounders \cite{kallus2018confounding, jung2018algorithmic}. 

\textbf{Off-policy evaluation with unobserved confounders}

Besides NKYB and KZ, most work in RL with unobserved confounders assumes that the causal effects are identified, i.e. assumptions are made about latent structure such that the true effect of interest can be recovered \cite{bennett2019policy, oberst2019counterfactual}. For POMDPs, \cite{tennenholtz2020off} analyze the bias for importance sampling in the presence of confounders, and give some conditions under which this bias can be corrected.

\section{Problem Setting and Notation}

\subsection{Markov Decision Processes}

Let $(\mathcal{X}, \mathcal{A}, P, R, \chi, \gamma)$ be an Markov decision process (MDP) where $\mathcal{X}$ is the set of states and $\mathcal{A}$ is the set of actions, which we assume are finite. Let $\mathcal{P}(S)$ denote all probability distributions on a set $S$. $P: \mathcal{X} \times \mathcal{A} \rightarrow \mathcal{P}(\mathcal{X})$ is the transition function, $R : \mathcal{X} \times \mathcal{A} \times \mathcal{X} \rightarrow \mathbb{R}$ is the reward function, $\chi \in \mathcal{P}(\mathcal{X})$ is the initial state distribution, and $\gamma \in [0,1)$ is the discount factor. A (stationary) policy $\pi : \mathcal{X} \rightarrow \mathcal{P}(\mathcal{A})$ assigns probabilities to each action given a state. We are interested in the expected value of policy $\pi$:
\[ V^{\pi}_T = \mathbb{E} \left[ \sum_{t=0}^{T-1} \gamma^t r_t \right], \]
where $x_0 \sim \chi$, $a_t \sim \pi(\cdot|x_t),~ x_{t+1} \sim P(\cdot| x_t,a_t),$ $r_t = R(x_t,a_t,x_{t+1})$ and $T \leq \infty$. 

\subsection{Confounded Off-policy Evaluation}

In this paper, we consider MDPs with unobserved confounding variables. Specifically, we assume the state space is partitioned into observed state $\mathcal{X}$ and unobserved state $\mathcal{U}$. The full-information MDP is $(\mathcal{X} \times \mathcal{U}, \mathcal{A}, P, R, \chi, \gamma)$.

In the confounded off-policy evaluation problem, we have access to a dataset $\mathcal{D}_{\pi_b} = \{\tau_i\}_{n=1}^N$, collected according to a stationary \emph{behavior} policy, $\pi_b : \mathcal{X} \times \mathcal{U} \rightarrow \mathcal{P}(\mathcal{A})$. Each $\tau_i = \{ (x_t^i, a_t^i, x_{t+1}^i, r_t^i) \}_{t=0}^{T-1}$ denotes an observed trajectory where $(x_0,u_0) \sim \chi$, $a_t \sim \pi_b(\cdot|x_t,u_t)$, $(x_{t+1}, u_{t+1}) \sim P(\cdot |x_t,u_t,a_t)$, and $r_t = R(x_t,a_t, x_{t+1})$. Note that while $R$ is only a function of the observed state, it can still rely on $u_t$ via $x_{t+1}$.

Our goal is to estimate the expected return $V^{\pi_e}_T$ for a stationary \emph{evaluation} policy, $\pi_e : \mathcal{X} \rightarrow \mathcal{P}(\mathcal{A})$, which does not depend on the unobserved state. 

\section{Two Types of Unobserved State}

We begin by making a distinction between unobserved states that are dependent over time, and unobserved states that are drawn iid each time step.

\textbf{Assumption 1} (IID Confounders). The unobserved state $u_t$ is drawn iid for all $t \geq 0$ and therefore the transition dynamics can be factored as:
\[P(x', u' | x, u, a) = P(x' | x, u, a) p(u') \]
This corresponds to the ``memoryless'' unobserved confounding assumption in KZ. Under Assumption 1, the marginal observed state transition probabilities are Markovian:
\[ P(x' | x, a) = \sum_{u \in \mathcal{U}} p(u) P(x' | x, u, a) \]
and the value of evaluation policy $\pi_e$ in the true MDP is equal to the value of $\pi_e$ in the marginal MDP, $(\mathcal{X}, \mathcal{A}, P, R, \chi, \gamma)$, where we abuse notation slightly to let $P$ and $\chi$ denote the corresponding marginal quantities over the observed state.

While we have reduced the problem to finding the value of $\pi_e$ in the marginal MDP, this value is \emph{not identified} given the dataset $\mathcal{D}_{\pi_b}$ because the unobserved state $u$ affects both the choice of action $\pi_b(a | x,u)$ and the transitions $P(x' | x,u,a)$. For example, any dataset collected under policy $\pi_b$ will be consistent with a set of many possible marginal transition probabilities $P(x' | x,a)$. However, standard OPE algorithms for MDPs can be adapted to this setting via some strategy to control for confounding. 

If the unobserved state is persistent, then the problem is no longer  a marginal MDP plus causal uncertainty. Consider the simplest such scenario where $u_0$ is drawn from some initial distribution and $u_t = u_0, \forall t$. In this setting, $P(x_{t+1} | x_t, a_t)$ is non-stationary in general and $P(x_{t+1} | x_t, a_t, ..., x_0, a_0)$ is not Markovian due to the dependence via $u$ induced by conditioning on $x$.  Therefore, the problem is a partially-observed MDP (POMDP).

For the POMDP case, even when $\pi_b(a | x, u) = \pi_b(a | x, u'), \forall a,x,u,u'$ (as in a randomized trial), many OPE algorithms are biased because the observed state and actions do not themselves constitute an MDP. A notable exception is IS methods. When $\pi_b(a | x, u) = \pi_b(a | x, u')$, the problem satisfies Assumption 1 in \cite{tennenholtz2020off} for POMDPs. 

When the behavior policy varies over $u$, the value is not identified and one must further adapt IS methods as in NKYB. However, as we will demonstrate, without Assumption 1 these bounds are too conservative for practical use - even when confounding is limited to a single time step. Therefore, in this paper, we develop lower bounds on the value of a policy given Assumption 1, and show that the bounds are far less sensitive to confounding. It is crucial to remember that Assumption 1 is not reasonable in some settings, especially medical ones, and given the substantial gap in performance, we suspect that new algorithms or sensitivity models need to be developed to make the persistent confounder case work in practice.

\section{Estimation with Unobserved Confounders}
\label{explicitbias}

\subsection{Bias due to Spurious Correlation}

Under Assumption 1, we can explicitly quantify the bias due to unobserved confounding. For comparison, we begin with a quantity that \emph{is} identified under confounding: the behavior policy conditional on the observed state. Consider the naive empirical estimate, $\hat\pi_b(a|x)$, for $\pi_b(a|x)$ given $\mathcal{D}_{\pi_b}$.

\begin{lem}
Under Assumption 1, $\hat \pi(a|x)$ is an unbiased estimator of $\pi_b(a|x)$. \begin{proof}\begin{align*}
    \mathbb{E}_{\mathcal{D}_{\pi_b}}[\hat\pi (a | x)] &= \sum_{u \in \mathcal{U}} p(u|x)\pi_b(a|x,u)\\ 
    &= \sum_{u \in \mathcal{U}} p(u)\pi_b(a|x,u)
    = \pi_b(a|x). \qedhere
\end{align*}
\end{proof}
\end{lem}

On the other hand, consider estimating the expectation of a function of $x,a,$ and $x'$, conditional on $x,a$, i.e. $m_f(x,a) \coloneqq \mathbb{E}[f(x,a,x') | x,a]$. Define the corresponding naive estimator, $\hat m_f(x,a)$ as above. 

\begin{prop}
Under Assumption 1 and given a function $f : \mathcal{X} \times \mathcal{A} \times \mathcal{X} \rightarrow \mathbb{R}$,
\begin{align*}
    &m_f(x,a) = \mathbb{E}_{\mathcal{D}_{\pi_b}}\left[\frac{\pi_b(a|x)}{\pi_b(a|x,u)}f(x,a,x') \Big| x,a\right].
\end{align*}
\begin{proof}[Proof sketch.]
Conditional on $x$ and $a$, the distribution of $u$ in $\mathcal{D}_{\pi_b}$ is $p(u|x,a)$ and
\[ p(u|x,a) = \frac{\pi_b(a|x,u)}{\pi_b(a|x)} p(u) \]
by Bayes rule. Then reweight accordingly. \qedhere
\end{proof}
\end{prop}
As an immediate corollary of Proposition 1, $\hat m_f(x,a)$ is not, in general, an unbiased estimator of $m_f(x,a)$. For a relevant example, let $f(x,a,x') = \mathbf{1}(x' = i)$ for some $i \in \mathcal{X}$. Then $m_f(x,a) = P(i | x, a)$, the marginal probability of transitioning to state $i$. Unless $\pi_b(a|x,u) = \pi_b(a|x,u')$ or $P(x'|x,a,u) = P(x'|x,a,u'), \forall u, u'$, the naive estimator of the transition probabilities is biased. Furthermore, since $\pi_b(a|x,u)$ is unobserved, the observed data is consistent with many possible $P(x'|x,a)$.

\subsection{Sensitivity Model}

While estimands like $P(x'|x,a)$ are not point-identified under Assumption 1, it is possible to give upper and lower bounds that are consistent with the observed data. However, without further assumptions these bounds are typically vacuous. Therefore, we follow the sensitivity analysis approach and specify limits on the impact of the unobserved state. The idea is that we will construct a worst-case estimate given a fixed level of confounding and study how the estimate changes as the degree of confounding is increased. 

We control the dependence of the behavior policy on the unobserved state via a parameter $\Gamma$. This is a popular technique in the causal inference literature, described in \cite{rosenbaum2002overt}. In particular, we follow \cite{tan2006distributional} and have $\Gamma$ bound the odds ratio between the unobserved behavior policy and the observed marginal behavior policy:

\textbf{Assumption 2} (Policy Confounding Bound). Given $\Gamma \geq 1$, for all $x \in \mathcal{X}$, $u \in \mathcal{U}$, and $a \in \mathcal{A}$:
\[ \frac{1}{\Gamma} \leq \left(\frac{\pi_b(a|x,u)}{1 - \pi_b(a|x,u)}\right) \Big/ \left(\frac{\pi_b(a|x)}{1 - \pi_b(a|x)}\right) \leq \Gamma \]
Note that Assumption 2 implies the bounds:
\[ \alpha(x,a) \leq \frac{\pi_b(a|x)}{\pi_b(a|x,u)} \leq \beta(x,a) \]
where
\begin{align*}
     \alpha(x,a) &\coloneqq \pi_b(a|x) + \frac{1}{\Gamma}(1-\pi_b(a|x))\\
     \beta(x,a) &\coloneqq \Gamma + \pi_b(a|x)(1-\Gamma)
\end{align*}

\section{Policy Evaluation with Confounders}

In this section, we will show how to compute worst-case value estimates. As long as Assumption 1 holds, by Proposition 1 we have an unbiased expression for regressing \emph{any} observed quantity $f$ against $x$ and $a$. This expression depends on the unknown probabilities $\pi_b(a|x,u)$ which we can bound using Assumption 2. By choosing different functions $f$, we can adapt most OPE direct methods as described in \cite{voloshin2019empirical}. We illustrate this procedure for Fitted Q Evaluation (FQE).

We begin with some notational details. We denote the state and state-action value functions for a policy $\pi$ and horizon $T$ as:
\begin{align*}
    V^\pi_T(x) &= \mathbb{E}\left[ \sum_{t=0}^{T-1} \gamma^t r_t \Big| x_0 = x \right]\\
    Q^\pi_T(x, a) &= \mathbb{E}\left[R(x,a,x') + V^\pi_{T-1}(x', u') \big| x, a \right]
\end{align*}
respectively. Throughout the rest of the paper, we will use the short-hand $g(x,\pi) \coloneqq \sum_{a \in \mathcal{A}} \pi(a|x)g(x,a)$. Denote the Bellman evaluation operator for a policy $\pi$ as $\mathcal{T}^\pi$, defined as:
\begin{align*} (\mathcal{T}^{\pi}g)(x,a) = \mathbb{E}\Big[r(x,a,x') + \gamma g(x',\pi) \Big| x, a\Big] \end{align*}
where $g$ is any function on $\mathcal{X} \times \mathcal{A}$. The state-action value function $Q^{\pi}_T$ can be computed by applying $\mathcal{T}^\pi$ to $Q_0 = 0$, T-times \cite{puterman2014markov}. Furthermore, $V^\pi_T(x) = Q^\pi_T(x,\pi)$, and the expected value is simply the average of the value function over the initial state distribution. Therefore, we can easily compute estimates of the expected value using $Q^\pi_T$.

\subsection{Confounded FQE}

FQE iteratively applies an empirical approximation of $\mathcal{T}^\pi$ to compute $Q^\pi_T$. Let $Q_0 = 0$ and let $\mathcal{H}$ be some function class. Given a dataset $\mathcal{D}_{\pi_b}$ and an evaluation policy $\pi_e$, FQE computes
\begin{align*}
    Q_k = \argmin_{h \in \mathcal{H}} \frac{1}{NT} \sum_{i=1}^N \sum_{t=0}^{T-1} (h(x_t^i, a_t^i) - y_t^i)^2\\
    \text{where } y_t^i = r(x_t^i,a_t^i,x_{t+1}^i) + \gamma Q_{k-1}(x_{t+1}^i, \pi_e).
\end{align*}
Essentially, regression with the class $\mathcal{H}$ approximates the conditional expectation of the function
\[ f(x,a,x') = r(x,a,x') + \gamma Q_{k-1}(x',\pi_e) \]
and $\mathcal{T}^{\pi_e}Q_{k-1}(x,a) = \mathbb{E}[f(x,a,x') | x,a]$. With unobserved confounding, regression using the data $\mathcal{D}_{\pi_b}$ no longer gives an unbiased estimate of $\mathcal{T}^{\pi_e}Q_{k-1}(x,a)$. Instead, we can apply Proposition 1 with the function $f$ defined above to get:
\[ \mathcal{T}^{\pi_e}Q_{k-1}(x,a) = \mathbb{E}_{\mathcal{D}_{\pi_b}}\left[\frac{\pi_b(a|x)}{\pi_b(a|x,u)}f(x,a,x') \Big| x,a\right]. \]
We can then use Assumption 2 to bound the unobserved $\pi_b(a|x,u)$. For example, we immediately get the following naive bound.
\begin{prop}
Let $y \coloneqq f(x,a,x')$. Under Assumptions 1 and 2, For all $x \in \mathcal{X}$ and $a \in \mathcal{A},$
\begin{align*} (\mathcal{T}&^{\pi_e}Q_{k-1})(x,a) \geq\\ &\mathbb{E}_{\mathcal{D}_{\pi_b}}\left[\left(\beta(x,a) \mathbf{1}(y < 0) + \alpha(x,a)\mathbf{1}(y \geq 0)\right) y \Big| x, a\right]. \end{align*}
\end{prop}

This naive bound is too conservative to use in practice, especially as the horizon grows. To get a better bound, we can solve an optimization problem over all possible values of $\pi_b(a|x,u)$ which are consistent with the observed data. Fix $x$ and $a$. Let $\pi_b(a|x)$ and $\hat P (x'|x,a)$ be the nominal behavior policy and nominal transition probabilities respectively. The basic unknown quantities are $p(u)$, $\pi_b(a|x,u)$, and $P(\cdot|x,u,a) \in \mathcal{P}(\mathcal{X}), \forall u$. We have the following observable implications:
\begin{lem}
Under Assumption 1, $\forall x \in \mathcal{X},a \in \mathcal{A},x' \in \mathcal{X}$,
\[ \sum_{u \in \mathcal{U}} p(u) \pi_b(a|x,u) = \pi (a|x), \text{ and} \]
\[\sum_{u \in \mathcal{U}} p(u) \pi_b(a|x,u) P(x'|x,u,a) = \pi(a|x) \hat P(x'|x,a). \]
\end{lem}

For a fixed $x$ and $a$, let $\mathcal{B}_{xa}$ be the set of possible $\pi_b(a|x,\cdot)$ such that Lemma 2 and Assumption 2 hold. Then:
\begin{align*} \mathcal{T}^{\pi_e}Q_{k-1}&(x,a) \geq\\ 
    &\min_{\pi_b(a|x,\cdot) \in \mathcal{B}_{xa}}\mathbb{E}_{\mathcal{D}_{\pi_b}}\left[\frac{\pi_b(a|x)}{\pi_b(a|x,u)}f(x,a,x') \Big| x,a\right]
\end{align*}
Unfortunately, when computing a regression in practice, this requires introducing a new optimization variable for the unknown values of $u$ for every data point. Instead we use a clever reparameterization to remove the dependence on $u$ that KZ introduced for MIS.

\subsection{Reparameterization}

Define
\[ g(x, a, x') \coloneqq \sum_{u \in \mathcal{U}} \left(\frac{p(u|x,a) P(x'|x,u,a)}{\hat P(x'|x,a)}\right) \frac{1}{\pi_b(a|x,u)} \]
and the corresponding set
\[ \Tilde{\mathcal{B}}_{xa} \coloneqq \{ g(x,a,\cdot) : \pi_b(a|x,u) \in \mathcal{B} _{xa}\}. \]
The idea is that $g(x,a,x')$ is equal to $1/\pi_b(a|x,u)$ convolved with an unknown density. Since both $\pi_b(a|x,u)$ and $p(u|x,a)P(x'|x,u,a)$ are unknown, optimizing over $\Tilde{\mathcal{B}}_{xa}$ is equivalent to optimizing over $\mathcal{B}_{xa}$ where we replace $\pi_b(a|x)/\pi_b(a|x,u)$ with $\pi_b(a|x)g(x,a,x')$. We have the following constraints:
\begin{lem}
Under Assumptions 1 and 2,\\ $\forall x \in \mathcal{X},a \in \mathcal{A},x' \in \mathcal{X}$,
\begin{align*}
    \alpha(x,a) \leq \pi_b(a|x)g(x,a,x') \leq \beta(x,a),
\end{align*}
and
\begin{align*}
    &\sum_{x' \in \mathcal{X}} \pi_b(a|x)g(x,a,x')\hat P(x'|x,a) = 1.
\end{align*}
\end{lem}

Now we are ready to state our confounded FQE bound:
\begin{thm} Under Assumptions 1 and 2,\\ $\forall x \in \mathcal{X}, a \in \mathcal{A}$,
\begin{align*} \mathcal{T}^{\pi_e}&Q_{k-1}(x,a) \geq\\ 
    &\min_{\pi_b(a|x,\cdot) \in \mathcal{B}_{xa}}\mathbb{E}_{\mathcal{D}_{\pi_b}}\left[\frac{\pi_b(a|x)}{\pi_b(a|x,u)}f(x,a,x') \Big| x,a\right]\\
    = &\min_{g(x,a,\cdot) \in \Tilde{\mathcal{B}}_{xa}}\mathbb{E}_{\mathcal{D}_{\pi_b}}\left[\pi_b(a|x)g(x,a,x')f(x,a,x') \Big| x,a\right]
\end{align*}
\end{thm}
For a given dataset $\mathcal{D}_{\pi_b}$, this bound can be computed with a simple linear program. Fix $x$ and $a$, and for shorthand, denote the naive estimates of the nominal behavior policy and nominal transition probabilities as $\hat \pi_{xa} \in [0,1]$ and $\hat P_{xa} \in [0,1]^{|\mathcal{X}|}$ respectively. The bound in Theorem 1 can be estimated by the following LP:
\begin{align*}
    \min_{w \in \mathbb{R}^{|\mathcal{X}|}}& c^T w \\
    &\text{ such that }\\
    &\hat \pi_{xa} + \frac{1}{\Gamma}(1-\hat \pi_{xa}) \preceq
    w \preceq \Gamma + \hat\pi_{xa}(1-\Gamma)\\
    &\text{ and } \hat P_{xa}^T w = 1,
\end{align*}
where $c(x')$ is the sample average of $r + \gamma Q_{k-1}(x',\pi_e)$ conditional on $x$ and $a$. Note that $\hat \pi_{xa}, \hat P_{xa},$ and $c$ are all observables estimated from the data, and $\Gamma$ is given. Only the vector $w$ is unknown. 

\textbf{Remark 1}. Theorem 1 gives a lower bound for a single application of $\mathcal{T}^{\pi_e}$. We get a lower bound on $V^{\pi_e}_k$ by applying $\mathcal{T}^{\pi_e}$ k-times and then averaging over the initial state distribution.

\textbf{Remark 2}.The reparameterized optimization problem in Theorem 1 can in principle be used when regressing a wide variety of functions $f$ against $x$ and $a$. This provides a blueprint for adapting other OPE methods that solve a regression problem.

\section{Sharper Bounds with Robust MDPs}
\label{model-based}
Unobserved variables create bias when they are correlated with \emph{both} the behavior policy \emph{and} the state transitions. The sensitivity model in Assumption 2 limits the correlation with the behavior policy. However, in the reparameterization strategy above, we combine our unknowns, $\pi_b(a|x,u)$ and $P(x'|x,u,a)$. Therefore, we cannot leverage any additional information that limits the correlation between $u$ and the transitions. Consider the extreme case, where $P(x'|x,u,a) = P(x'|x,a), \forall u$. In this case, naive OPE estimates will be unbiased even if $\Gamma$ in Assumption 2 is large. While in observational studies, it is not possible to rule out all correlation between unobservables and the dynamics, we might be able to use domain knowledge on causal mechanisms to restrict the feasible transitions. 

One branch of the sensitivity analysis literature, exemplified by \cite{rosenbaum1983assessing}, suggests using three sensitivity parameters. First, a bound on the correlation between the unobserved confounder and the treatment. Second, a bound on the correlation between the unobserved confounder and the outcome. Third, a parameter representing the distribution of the unobserved confounder. \cite{rosenbaum1983assessing} presents the case where $u$ is a binary variable. However, \cite{ding2016sensitivity} show that for worst-case bounds, this is without loss of generality. Therefore, we assume that $\mathcal{U} = \{0,1\}$.

Assumption 2 bounds the impact of $u$ on $\pi_b$. Following \cite{rosenbaum1983assessing}, we now introduce two additional parameters:

\textbf{Assumption 3} (Transition Confounding Bound). Given $\Delta \geq 1$, for all $x \in \mathcal{X}$, $a \in \mathcal{A}$, $x' \in \mathcal{X}$, and $u \in \mathcal{U}$:
\[ \frac{1}{\Delta} \leq \Bigg( \frac{P(x'|x,u,a)}{1-P(x'|x,u,a)} \Bigg) \Big/ \Bigg(  \frac{\hat P(x'|x,a)}{1-\hat P(x'|x,a)} \Bigg) \leq \Delta \]

\textbf{Assumption 4}. Given a fixed $p \in [0,1], p(u=1) = p$.

For any tuple of sensitivity parameters, $(\Gamma, \Delta, p)$, we will give worst-case bounds on the value function $V^{\pi_e}_T(x)$ using a model-based approach. Each $(\Gamma, \Delta, p)$ has a corresponding set of possible transition probabilities under Assumptions 2, 3, and 4, such that the observable implications in Lemma 2 hold. Finding the worst-case value given an uncertainty set for the dynamics has been extensively explored in the Robust MDP literature \cite{nilim2005robust}. The standard approach is to separate the uncertainty over the state-action pairs, assuming that the uncertainty sets across $x,a$ pairs are not linked. In our problem, this assumption is violated because of the requirement that $\pi_b(\cdot|x,u)$ is a probability distribution. In the language of robust MDPs, our problem is ``s-rectangular'' instead of ``s,a-rectangular''. 

Fortunately, s-rectangular MDPs can also be solved efficiently \cite{wiesemann2013robust}. Let $\mathcal{G}_x$ denote the set of feasible transition probabilities for a fixed $x$. Let $P_x \in \mathbb{R}^{|\mathcal{A}| \times |\mathcal{X}|}$ be the matrix whose rows are $P(\cdot|x,a)$ for each $a$. Instead of the state-action value function, we iteratively solve for worst-case estimates of the value function:
\[ V_k(x) = \min_{P_x \in \mathcal{G}_x} \pi_e(\cdot|x)^T P_x y \]
where $y = (\sum_{a\in\mathcal{A}} \pi_e(a|x) R(x,a,\cdot)) + \gamma V_{k-1}(\cdot)$. When optimizing over the unknown quantities $P(x'|x,u,a)$ and $\pi_b(a|x,u)$ for all $x'$, $a$, and $u$, this problem has a linear objective with linear and bilinear equality constraints, so it can be easily solved. We estimate $V_T^{\pi_e}$ by letting $V_0 = 0$, then solving the above minimization problem $T$-times. As we will show in our evaluation, for all values of the parameters $(\Gamma, \Delta, p)$, the s-rectangular robust MDP formulation provides sharper bounds than the linear program corresponding to Theorem 1. 

\section{Evaluation}

We use the benchmarks from OPE-Tools \cite{voloshin2019empirical} for evaluation. In particular, we adapt their three discrete environments, Graph, Discrete MC, and Gridworld, together with a small toy problem. Note that the data generating processes do not strictly need to be confounded. Our methods bound the worst possible confounded MDP that \emph{could} have generated the data. Therefore, the two relevant, observable reference points are the value of the behavior policy and the nominal value of the evaluation policy. Nonetheless, for completeness we augment the environments with unobserved confounding variables. Our approach takes an existing behavior policy and transition matrix, and adds an additional state variable $u$ which induces a correlation between the policy and transitions based on either the rewards or the optimal value function.

For each environment, we choose a behavior policy $\pi_b$ and evaluation policy $\pi_e$ such that the value of $\pi_e$ without confounding is greater than the value of $\pi_b$. This way, it is possible to find which level of confounding makes it impossible to guarantee that $\pi_e$ is superior to $\pi_b$. Furthermore, the impact of confounding can be compared relative to the difference in values between the two policies. See Table $\ref{tab:env_desc}$ for a summary of the four test environments and the Appendix for full details.

\begin{table*}
\centering
\begin{tabular}{ | c | c | c | c | c | c | c |}
 \hline
 Environment & Horizon & States & Actions & $V^{\pi_b}_T$ & $V^{\pi_e}_T$ & Sparse Rewards? \\
 \hline\hline
 toy & 5 & 3 & 2 & 0.3397 & 0.4990 & No \\  
 \hline
  ope-graph & 4 & 8 & 2 & -0.1786 & 0.7174 & No \\  
  \hline
   ope-mc & 20 & 22 & 2 & -18.1890 & -15.7381 & Yes \\  
   \hline
    ope-gridworld & 8 & 16 & 4 & -0.4994 & -0.3569 & No \\  
    \hline
\end{tabular}
\caption{\label{tab:env_desc} Characteristics of the four test environments.}
\end{table*}

\subsection{Lower Bounds with Confounding}

\begin{figure}[ht]
    \centering
    \includegraphics[width=0.5\textwidth]{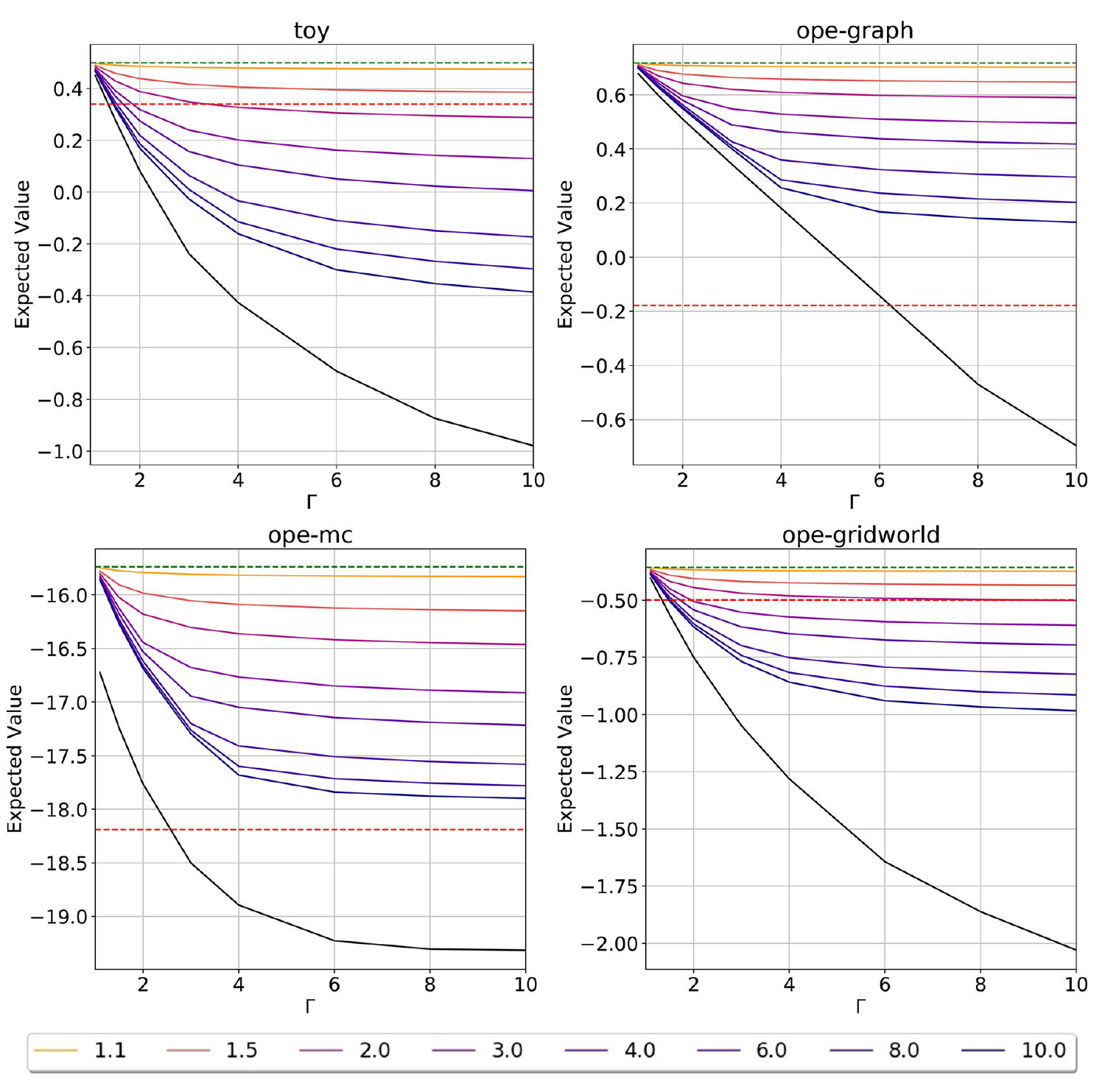}
    \caption{Lower bounds on the expected value of $\pi_e$. For reference, in each environment, we plot the value of $\pi_e$ without confounding (the dotted line at the top) and the value of $\pi_b$ (the dotted line below). The black line at the bottom is the confounded-FQE bound. Each other line corresponds to a robust MDP bound for a single value of the transition confounding parameter $\Delta$, with light to dark lines going from 1.1 to 10.}
    \label{fig:gamma_delta}
\end{figure}

For our first experiment, we collect trajectories from each of the four environments using their respective behavior policies. For each environment, we collect 30,000/horizon trajectories, keeping the number of data points the same across environments. Then, we compute our confounded FQE and robust MDP lower bounds for values of $\Gamma$ and $\Delta$ ranging between 1.1 (barely confounded) and 10 (highly confounded). For the robust MDP bounds, we fix the parameter $p = 0.5$, i.e. each period the unobserved state is equally likely to be $u=0$ or $u=1$. The robust MDP bounds are not very sensitive to this parameter and this choice doesn't impact the qualitative results, although corroborating results are in the Appendix. Our lower bounds for the four environments are plotted in Figure \ref{fig:gamma_delta}. 

The confounded FQE bounds are the black curve at the bottom of each plot. Without any additional restrictions of the transition dynamics, these bounds degrade the quickest as $\Gamma$ increases. This curve intersects the value of $\pi_b$ at $\Gamma = 6$ for ope-graph, and $\Gamma < 3$ for the remaining environments. Qualitatively, this means strong requirements on confounding are required for the FQE bounds to guarantee that the evaluation policy is better than the behavior policy. Compare this, for example, to the other curves in ope-graph and ope-mc which are greater than $V^{\pi_b}$ for all values of $\Gamma$. 

The curves above the confounded FQE curve correspond to our robust MDP bounds. In all cases as $\Delta$ grows, the corresponding lower bounds get worse. As mentioned previously, for ope-graph and ope-mc, any value of $\Delta$ guarantees that $V^{\pi_e} > V^{\pi_b}$. For toy and ope-gridworld, consider the $\Delta = 2$ curve, which is third from the top. For the toy environment, assuming $\Delta = 2$ substantially increases the $\Gamma$ at which the curve crosses the dotted $\pi_b$ line compared to the FQE curve. For ope-gridworld, the $\Delta=2$ curve lies above $V^{\pi_b}$ for all $\Gamma$. These examples highlight the qualitative and quantitative importance of limiting the degree of confounding on the transition probabilities.

\subsection{Tightness}

Our confounded FQE and robust MDP methods provide lower bounds on the expected value subject to their respective sensitivity models. A natural question is: how far are these bounds from the infimum over all full-information MDPs consistent with the observed data, subject to the given sensitivity model? We split our analysis of tightness into two parts, the single-step case and the multi-step case. 

A single iteration of our bounds requires solving a minimization problem. The tightest possible bound on $V_{T}^\pi$ is the minimum over all valid full-information MDPs. But our robust MDP solution produces candidate transition probabilities $P(x'|x,u,a)$ and behavior policy $\pi_b(a|x,u)$ corresponding to some valid full-information MDP. Therefore, since it is a lower bound, it must achieve the true minimum and so a single iteration of the robust MDP approach is tight. 

\begin{figure}[ht]
    \centering
    \includegraphics[width=0.5\textwidth]{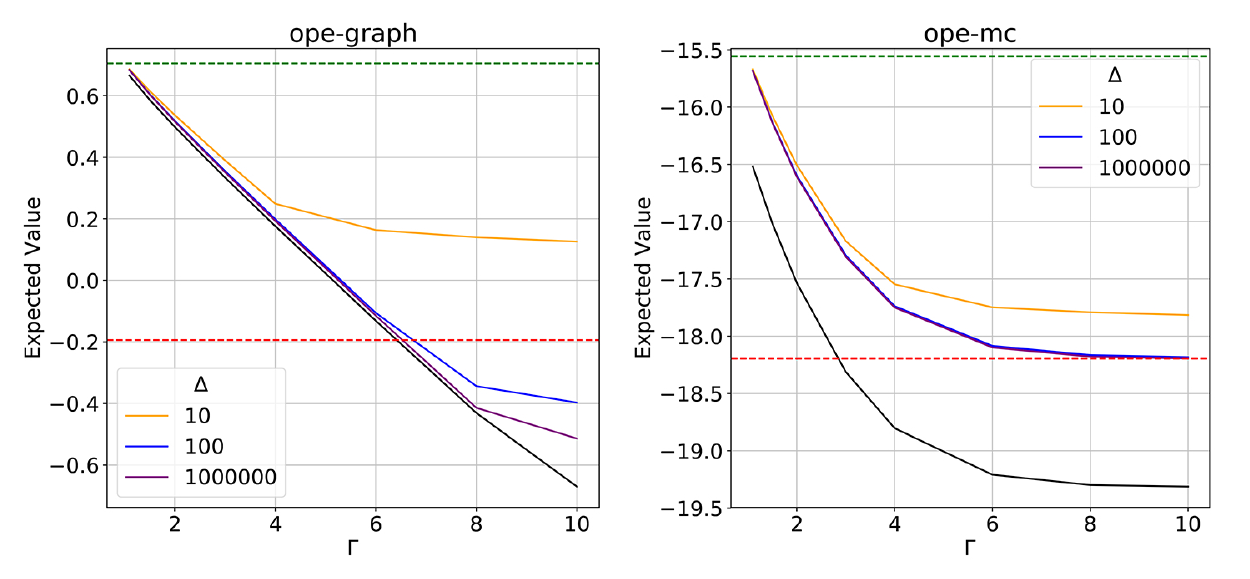}
    \caption{Lower bounds on the expected value as $\Delta$ grows large. The black line at the bottom is the confounded FQE bound. The upper dashed line is value of $\pi_e$ with no confounding. The lower dashed line is the value of $\pi_b$. }
    \label{fig:sa_rect}
\end{figure}

On the other hand, our confounded FQE bound solves a minimization problem separately for each state-action pair without enforcing that $\pi_b(\cdot|x,u)$ be a density across actions. We quantify the impact on performance by comparing the FQE bound to our robust MDP bound as $\Delta$ goes to infinity. We present results for the ope-graph and ope-mc environments in Figure \ref{fig:sa_rect}. The qualitative findings for the other environments are similar so we defer these results to the Appendix.

For the ope-graph environment, the gap between the FQE bound (the black line at the bottom) and the robust MDP bounds for large $\Delta$ are negligible until $\Gamma \geq 8$, at which point the gap grows. For the ope-mc environment, the gap begins substantial and grows slightly larger as $\Gamma$ grows. For this particular environment, the robust MDP lower bounds always guarantee that the evaluation policy is at least as good as the behavior policy. However, the FQE lower bound can only provide this same guarantee for $\Gamma < 3$. Therefore, it appears that enforcing the density constraint across actions can matter in practice, so for cases where we do not wish to make any assumptions on the transitions, we prefer our robust MDP bounds with very large values of $\Delta$. 

When confounding occurs in more than one time step, our robust MDP bound is computed iteratively with different minimization problems solved at each time step. The candidate transitions and behavior policy that correspond to each minima may differ, so the lower bounds are potentially loose. Theoretically, the looseness of our bound is characterized by Theorem 4 of \cite{nilim2005robust}. In particular, as the horizon goes to infinity, our lower bound converges to the best possible lower bound - the rate of convergence can be found in the proof of the theorem. 

To test this empirically, we use the full-information transitions and behavior policy from the final iteration of our robust MDP method as a candidate. Because the candidate MDP is consistent with the observed data subject to the sensitivity model, if the value of this MDP matches our lower bound, than our lower bound must be tight. For the toy, ope-graph, and ope-mc environments, we use the same experimental setup as we did for the results in Figure \ref{fig:gamma_delta}. The gap between the candidate MDP value and our lower bounds are reported in Table \ref{tab:horizon_tightness}. For these environments, the value of the candidate MDP differs by less than $10^{-8}$ from our lower bound. For the ope-gridworld environment, we find our lower bound is not tight at small horizons, so we ran experiments with a short, medium, and long horizon. As predicted by the theory, the bound improves for large $T$ as value iteration approaches its fixed point. 

\begin{table}[t]
\centering
\begin{tabular}{ | c | c | c | }
 \hline
 env & $\Gamma$ = 2, $\Delta$ = 2 & $\Gamma$ = 10, $\Delta$ = 10\\
 \hline\hline
 toy & $<$ 1e-8 & $<$ 1e-8 \\  
 \hline
 ope-graph & $<$ 1e-8 & $<$ 1e-8\\
 \hline
 ope-mc & $<$ 1e-8 & $<$ 1e-8\\
 \hline
 ope-gridworld T=28 & 2.03e-3 & 3.06e-2\\
 \hline
 ope-gridworld T=208 & 4.75e-3 & 2.87e-2\\
 \hline
 ope-gridworld T=508 & 2.65e-5 & 2.97e-4\\
 \hline
\end{tabular}
\caption{\label{tab:horizon_tightness} The difference between our robust MDP bound and the value of $\pi_e$ in the candidate MDP defined by the transition probabilities from the last iteration of our bound. The first three environments use the default horizons given in Table \ref{tab:env_desc}.}
\end{table}

\subsection{Assumption 1 and Comparison with NKYB}

Assumption 1 - that the unobserved state is drawn iid each period - is crucial to the quality of the bounds above. We demonstrate this by comparing our bounds to those in NKYB, which do not assume iid confounders. In order to compare to NKYB, we have to alter the experimental setup above in two ways. First, NKYB only supports confounding that occurs in a single time step. The initial time step is confounded, but for the remainder of the horizon, the behavior policy only uses the observed state. We compute the analogue for our robust MDP algorithm by computing $T-1$ iterations of unconfounded value iteration followed by a single iteration of our lower bound. 

Second, NKYB uses a similar but more restrictive sensitivity model. Our sensitivity parameter restricts the odds ratio between the confounded policy for a given value of $u$ and the policy averaged over all $u$. Their sensitivity parameter restricts the odds ratio for the confounded policy between any values of $u$, which grows roughly like the square of ours. For this comparison, we can calculate the true sensitivity parameters for each confounded environment under the  different sensitivity models. We provide a performance comparison using the true sensitivity parameters for each environment in Table \ref{tab:nkyb_compare}. Even with confounding restricted to a single time step, the NKYB bounds, which do not assume iid confounders, are enormously conservative. 

This is a key result. Even for a single time-step, policy evaluation is highly sensitive to persistent unobserved variables. The ability of our robust MDP bounds to guarantee improvement over the behavior policy in Figure \ref{fig:gamma_delta}, even over longer horizons, depends crucially on our Assumption 1. In turn, this highlights the fact that off-policy evaluation with confounding in settings where Assumption 1 fails is far more difficult and requires a different algorithmic approach. As mentioned in the introduction, while iid confounders are feasible in certain settings - like unobserved oil supply shocks for macroeconomic policy - Assumption 1 is \emph{not} reasonable for many applications, especially in medicine.

The results in Table \ref{tab:nkyb_compare} might hinge on the different sensitivity models, so we perform a robustness check which uses identical values of $\Gamma$ and which should therefore be very favorable for the NKYB bounds. The toy and ope-graph NKYB bounds improve, but the ope-mc and ope-gridworld bounds remain unusable - see the Appendix for details. 

\begin{table}
\centering
\begin{tabular}{ | c | c | c | c | }
 \hline
 env & Nominal & NKYB & Ours\\
 \hline\hline
 toy & 0.5189 & 0.0436 & 0.25372 \\  
 \hline
 ope-graph & 0.7008 & 0.0280 & 0.3994\\
 \hline
 ope-mc & -15.6941 & -64.5040 & -15.9647\\
 \hline
 ope-gridworld & -0.3588 & -2.3914 & -0.4112\\
 \hline
\end{tabular}
\caption{\label{tab:nkyb_compare} The value of $\pi_e$ without confounding and the corresponding lower bounds from NKYB and our robust MDP procedure. For each bound and each environment, we use the true parameter value for the respective sensitivity models.}
\end{table}

\subsection{Horizon and Comparison with KZ}

Many of the details above depend on the horizon. For example, our robust MDP bounds become tight as the horizon increases and NKYB restricts confounding to a single time-step. Therefore, in this section we assess how our lower bounds change as the horizon increases. This also provides a convenient setting to compare with the infinite horizon bounds in KZ.

Comparing with KZ requires modification of our initial experimental setup. We use the ope-graph and ope-gridworld environments. In order to generate a non-trivial steady-state distribution, we remove the terminating states and alter the transition probabilities accordingly. Furthermore, to match KZ's approach, we modify the rewards to only depend on the current state. We then calculate our bounds for 1 to 200 time steps. For both environments, $T=200$ is long enough to spend a majority of the time close to steady-state. We also adopt a discount rate of $\gamma = 0.95$ so that $T=200$ is well beyond the effective horizon. We produce bounds for $\Gamma = 1.5, 2,$ and $10$ using our robust MDP method with $\Delta$ set to 1,000,000.

\begin{figure}[ht]
    \centering
    \includegraphics[width=0.5\textwidth]{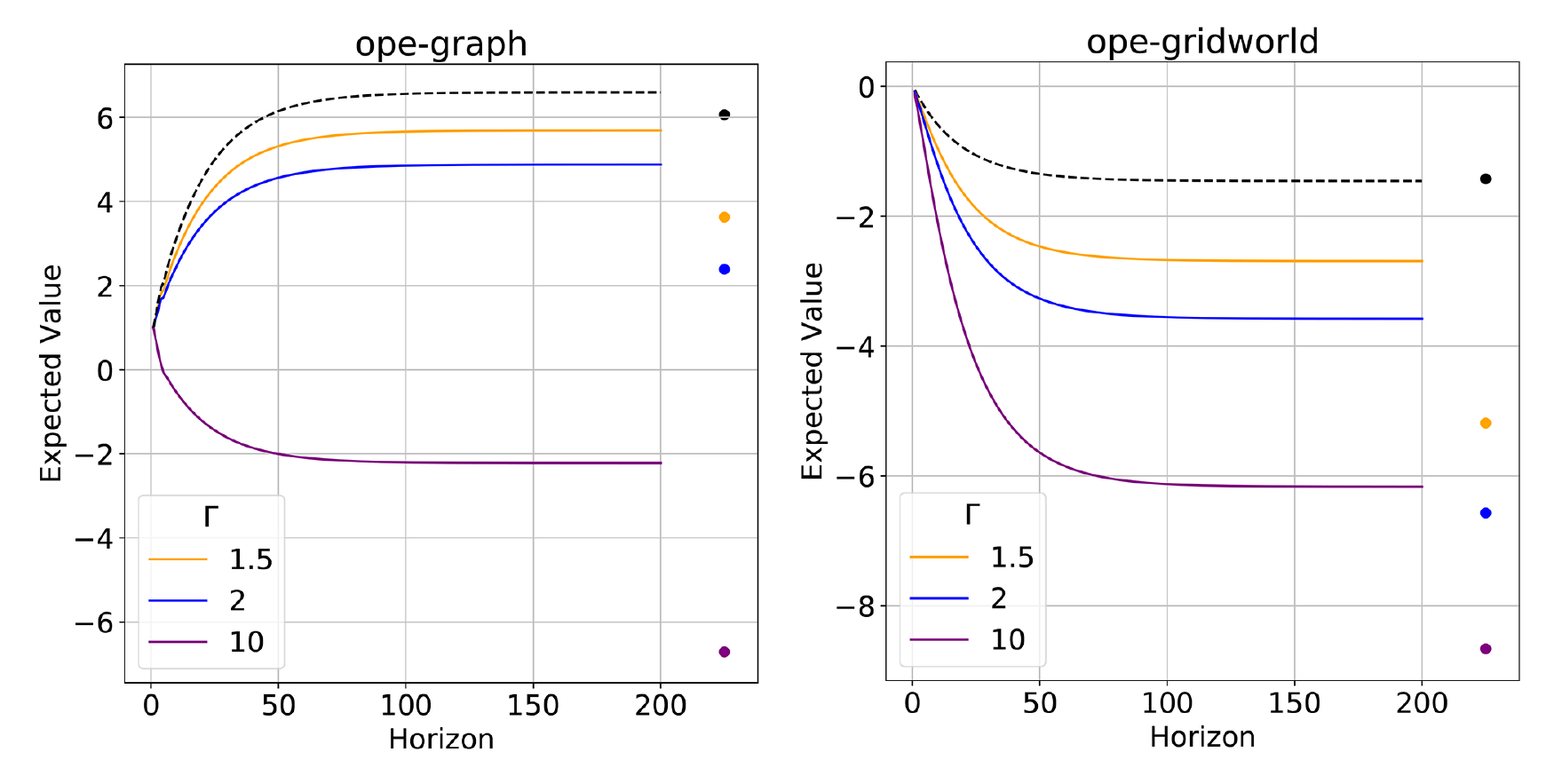}
    \caption{Robust MDP lower bounds as the horizon grows. The dotted curve is the nominal value of $\pi_e$. The dots on the right are KZ's infinite horizon bounds.}
    \label{fig:infinite_horizon}
\end{figure}

Since we use the same marginal sensitivity model, we can use KZ's method to calculate infinite horizon bounds for the same values of $\Gamma$. Their method computes bounds on the long-run average value, i.e. the expectation of the rewards with respect to the steady-state distribution, instead of the discounted value. Therefore we use the discounted sum of rewards as the per-state reward for KZ's method. The results are plotted in Figure \ref{fig:infinite_horizon}. The dotted black curve at the top is the value of $\pi_e$ without confounding at each horizon. The curves below are the lower bounds for $\Gamma = 1.5, 2,$ and $10$ respectively. The dots on the far right are the corresponding KZ infinite horizon bounds. In all cases, the gap between our bounds and the unconfounded value grow at the horizon increase. This is not because our bounds are loose - as value iteration reaches its fixed point, our bounds are provably tight as mentioned - but because confounding over many time periods is a more difficult problem. This phenomenon is especially pronounced for $\Gamma = 10$: at long horizons,  a smaller value of the sensitivity parameters becomes much more valuable. 

The infinite horizon bounds follow roughly the same qualitative behavior as ours but are much looser. This is presumably due to the fact that the long-run average of discounted rewards is a different estimand than the average discounted sum of rewards. With no confounding, the difference is small (compared the uppermost line and uppermost dot). But as the level of confounding increases, the long-run average becomes more sensitive. The magnitude of the difference is surprising and perhaps worth studying in future work.  

\section{Conclusion}

To summarize: our first key contribution is to develop a method for computing finite horizon lower bounds for policy evaluation with unobserved confounders that are drawn iid each period. We find that our model-based robust MDP approach can give substantially sharper bounds by leveraging assumptions about the transition probabilities. To be clear on this point: the argument is not that a plug-in estimator using a model of the dynamics is inherently more efficient. When using observational data to estimate a dynamic causal effect, understanding the dynamics of the system and the causal mechanisms are critically important. Quantitatively, we illustrate this by showing that sharp partial identification of the value of a policy requires restricting the set of possible transition probabilities. In practice, such an approach relies on domain-expertise. Practitioners must have enough mechanistic understanding of the dynamics that they are able to specify bounds, $\Delta$, on potential confounding in order to get a reasonable estimate of the expected value.

Our second key contribution is to demonstrate that policy evaluation is far more challenging when there are persistent unobserved confounders. This is responsible for the substantial performance gap between our bounds and those in NKYB. These results are especially relevant for medical applications where unobserved variables are likely to be persistent. For example, any patient variable that may not be recorded, but doesn't change between treatment choices like socio-economic status or undocumented chronic illness. Work published after this paper was completed \cite{kwon2021rl} has taken an initial step to tackle this setting without confounding. An important next step will be to achieve similar results in the observational causal setting. 



\bibliography{main}
\bibliographystyle{icml2021}

\end{document}